\documentclass[letterpaper, 10 pt, conference]{IEEEtran}
\IEEEoverridecommandlockouts

\usepackage{cite}
\usepackage{amsmath,amssymb,amsfonts}
\usepackage{algorithmic}
\usepackage{graphicx}
\usepackage{textcomp}
\usepackage{xcolor}
\usepackage{multirow}
\usepackage{cleveref}
\begin{document}

\title{Efficient Traffic Prediction at Scale: A Systematic Study of STGCN Architectural Depth

\thanks{\textcopyright~2026 IEEE. Personal use of this material is permitted. Permission from IEEE must be obtained for all other uses, in any current or future media, including reprinting/republishing this material for advertising or promotional purposes, creating new collective works, for resale or redistribution to servers or lists, or reuse of any copyrighted component of this work in other works.}
\thanks{This work was supported by the International Graduate School of Science and Engineering (IGSSE) of the Technical University of Munich (TUM) through the MINDMAP project.}
\thanks{GitHub repository: https://github.com/tum-tse/stgcn-lite}

}
\author{%
\null\hfill
\begin{minipage}[t]{0.3\textwidth}\centering
Soban Nasir Lone\\
\textit{Technical University of Munich}\\
Munich, Germany\\
soban.lone@tum.de
\end{minipage}\hfill
\begin{minipage}[t]{0.3\textwidth}\centering
Mohamed Abouelela\\
\textit{Technical University of Munich}\\
Munich, Germany\\
mohamed.abouelela@tum.de
\end{minipage}\hfill
\begin{minipage}[t]{0.3\textwidth}\centering
Taeyoung Yu\\
\textit{The University of Queensland}\\
Brisbane, Australia\\
taeyoung.yu@uq.edu.au
\end{minipage}\hfill\null
\\[1.5em]
\null\hfill
\begin{minipage}[t]{0.3\textwidth}\centering
Jiwon Kim\\
\textit{The University of Queensland}\\
Brisbane, Australia\\
jiwon.kim@uq.edu.au
\end{minipage}\hspace{0.04\textwidth}%
\begin{minipage}[t]{0.3\textwidth}\centering
Constantinos Antoniou\\
\textit{Technical University of Munich}\\
Munich, Germany\\
c.antoniou@tum.de
\end{minipage}\hfill\null
}

\maketitle

\begin{abstract}

Spatio-temporal graph neural networks (STGNNs) have become the dominant approach 
for traffic prediction, yet their computational requirements pose challenges for 
practical deployment in intelligent transportation systems (ITS). While recent work 
has proposed efficient alternatives to STGNNs, a fundamental question remains 
unexplored: are these architectures themselves over-parameterised? We examine this 
question using the Spatio-Temporal Graph Convolutional Network (STGCN), one of the most widely adopted models in this domain. Through 
systematic experiments across four diverse traffic datasets, we compare 1-block, 
2-block (standard), and 3-block STGCN variants. Our findings reveal that the 
single-block architecture achieves optimal performance for short-term prediction 
(10 mins) on three of four datasets, while incurring only marginal degradation 
($\leq$1.8\% relative error) at longer horizons. Crucially, the 2-block variant 
incurs 61\% higher CPU inference latency and 37\% lower throughput relative to 
1-block -- substantial overhead for resource-constrained ITS deployment. The 
3-block architecture offers no favourable tradeoff, more than doubling 
computational cost for $<$0.5\% relative improvement. These results suggest that 
the default 2-block STGCN may be over-parameterised for many applications, with 
implications for both practitioners deploying traffic prediction systems and 
researchers benchmarking efficiency-focused methods.
\end{abstract}

\begin{keywords}
Traffic prediction, spatio-temporal graph neural networks, STGCN, model efficiency, intelligent transportation systems
\end{keywords}

\section{Introduction}

Accurate traffic prediction is fundamental to intelligent transportation systems (ITS), enabling applications from adaptive signal control to route guidance and congestion management. The past decade has seen spatio-temporal graph neural networks (STGNNs) emerge as the dominant paradigm for this task, with models such as Spatio-Temporal Graph Convolutional Network (STGCN)~\cite{yu2018spatio}, Diffusion Convolutional Recurrent Neural Network (DCRNN)~\cite{li2018diffusion}, and Graph WaveNet~\cite{wu2019graph} consistently achieving state-of-the-art performance across benchmark datasets \cite{GraphNeuralNetworkjiang2022}.

However, the computational demands of STGNNs present practical challenges for real-world deployment. Traffic management systems often operate under strict real-time requirements -- adaptive signal control systems adjust signal timing parameters `instantaneously' or on an `ongoing basis' to accommodate traffic variability \cite{fhwa2017}, while city-wide prediction must scale to thousands of intersections \cite{zhan2016citywide}. Many deployed systems rely on resource-constrained infrastructure due to cost, power consumption, and hardware availability constraints, where inference efficiency becomes critical \cite{RealTimeUrbankhattab2025}.

This tension between accuracy and efficiency has motivated a growing body of work on lightweight traffic prediction models. Recent studies have questioned whether the complexity of graph neural networks is necessary at all: SimST~\cite{liu2023we} demonstrated that simple temporal models can achieve competitive performance with 39$\times$ higher throughput, while STGformer~\cite{jiang2024stgformer} and LightST~\cite{Zhang_Gao_Wang_Yiu_Yin_2025} propose efficient alternatives through architectural innovations and knowledge distillation. These works consistently benchmark against STGCN as the representative baseline for spatio-temporal graph approaches.

Yet a fundamental question remains unexplored: \emph{is STGCN itself over-parameterised?} The original STGCN architecture~\cite{yu2018spatio} employs two stacked spatio-temporal blocks, a design choice that has been adopted without systematic justification. Subsequent work has largely inherited this default, treating the 2-block configuration as the standard STGCN. If simpler variants perform comparably, this has direct implications for both deployment efficiency and the validity of using 2-block STGCN as an efficiency baseline.

In this paper, we conduct a systematic investigation of STGCN depth across four diverse traffic datasets spanning the United States (US) highway and Chinese urban networks. We compare 1-block, 2-block, and 3-block variants, measuring both predictive performance and computational cost -- including CPU inference time, which is often overlooked but critical for practical deployment. The key highlights of our findings can be visualised in Fig. 1.

Our contributions are as follows:
\begin{itemize}
    \item We demonstrate that 1-block STGCN matches or outperforms the standard 
    2-block variant for short-term prediction (10 mins) on three of four datasets, 
    with maximum degradation of 1.8\% MAE at longer horizons, despite incurring 
    38\% lower CPU inference latency and 60\% higher throughput.
    \item We show that 3-block STGCN offers no favourable tradeoff, more than 
    doubling computational cost for negligible ($<$0.5\%) or negative performance 
    impact.
    \item We suggest that depth conventions in STGNNs more broadly may be 
    inherited rather than justified, warranting systematic ablation across 
    architectures with similar stacked designs.
    \item We propose that efficiency comparisons in the STGNN literature adopt 
    the minimal competitive architecture as baseline, rather than community 
    defaults that may inflate reported gains.
\end{itemize}

\begin{figure}
    \centering
    \includegraphics{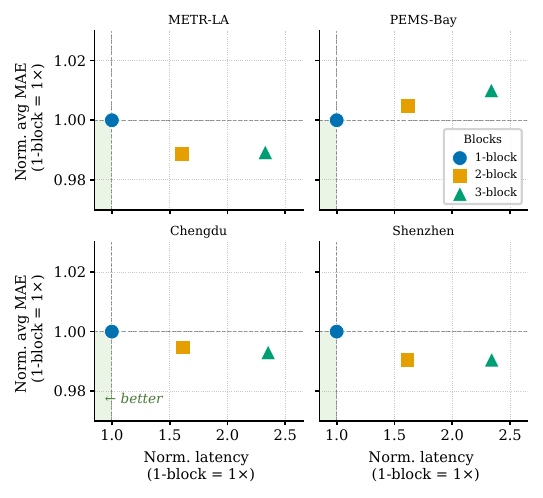}
    \caption{Efficiency–accuracy tradeoff for STGCN depth variants, normalised to the 1-block baseline. Lower-left is optimal (shaded). The 1-block variant is Pareto-optimal on PEMS-Bay; on remaining datasets, deeper variants yield marginal accuracy gains ($\leq$1.8\% MAE) at disproportionate latency cost (1.6–2.3×).}
    \label{fig:tradeoff}
\end{figure}
\section{Related Work}

\subsection{Spatio-Temporal Graph Neural Networks}

STGNNs jointly model spatial dependencies through graph convolutions and temporal dynamics through sequential architectures. STGCN~\cite{yu2018spatio} introduced the paradigm of stacking spatio-temporal convolutional blocks, using graph convolutions for spatial modelling and gated temporal convolutions for capturing temporal patterns. DCRNN~\cite{li2018diffusion} combined diffusion convolutions with recurrent units, while Graph WaveNet~\cite{wu2019graph} introduced adaptive adjacency matrices with dilated causal convolutions.

The original STGCN paper proposed a 2-block architecture, and subsequent work has begun to shed light on how depth influences model performance. For instance, the H-STGCN paper~\cite{dai2020hstgcn} observed that ``one block is found sufficient to achieve a similar level of accuracy'' on their dataset, suggesting that shallower variants may be effective in certain settings. Building on these initial observations, our work presents a systematic analysis of architectural depth in STGCN across multiple datasets.

\subsection{Efficient Traffic Prediction}

Recent work has increasingly focused on the efficiency-accuracy tradeoff in traffic prediction. STGformer \cite{jiang2024stgformer} analysed the error versus floating point operations per second (FLOPs) across architectures, noting that Transformer-based approaches can be less efficient than well-designed Graph Convolution Network (GCN) models. They also test their proposed method on larger benchmark datasets \cite{liu2023largest}. SimST questioned whether graph neural networks are necessary at all, showing that simple temporal models achieve competitive results with dramatically lower computational cost \cite{liu2023we}. LightST proposed knowledge distillation from complex teachers to lightweight students \cite{Zhang_Gao_Wang_Yiu_Yin_2025}.

These works share a common methodological choice: they benchmark against the standard 2-block STGCN as the representative GCN-based baseline. Our findings suggest that 1-block STGCN may be a more appropriate efficiency baseline, providing a fairer comparison point and better reflecting the minimal architecture required for competitive performance.

\section{Methodology}

\subsection{STGCN Architecture}

The STGCN model processes traffic data represented as a graph signal $\mathcal{X} \in \mathbb{R}^{N \times T \times C}$, where $N$ is the number of nodes (sensors), $T$ is the number of timesteps, and $C$ is the number of features. The graph structure is encoded in an adjacency matrix $A \in \mathbb{R}^{N \times N}$ derived from road network connectivity. The adjacency matrices are defined either by the distance between sensors, normalised by a thresholded Gaussian kernel \cite{yu2018spatio}, or by network connectivity (1 if connected, 0 if not), following similar practices in the literature \cite{STG4TrafficSurveyBenchmarkluo2024}. Dataset-related adjacency details are presented in \Cref{tab:datasets}.

The core building block is the ST-Conv block, which consists of:
\begin{enumerate}
    \item \textbf{Temporal convolution}: Gated temporal convolution capturing local temporal dependencies.
    \item \textbf{Spatial convolution}: Chebyshev graph convolution modelling spatial relationships across the network.  
    \item \textbf{Temporal convolution}: Second gated temporal convolution for additional temporal processing.
\end{enumerate}

The standard STGCN stacks two such blocks followed by an output layer. We investigate three variants:
\begin{itemize}
    \item \textbf{STGCN-1B}: Single ST-Conv block + output layer.
    \item \textbf{STGCN-2B}: Two ST-Conv blocks + output layer (original design).
    \item \textbf{STGCN-3B}: Three ST-Conv blocks + output layer.
\end{itemize}

\subsection{Datasets}

We evaluate on four traffic prediction benchmarks spanning different network sizes and geographical contexts, as presented in \Cref{tab:datasets}.

\begin{table}[h]
\centering
\caption{Dataset characteristics}
\label{tab:datasets}
\begin{tabular}{lcccc}
\hline
\textbf{Dataset} & \textbf{Nodes} & \textbf{Timesteps} & \textbf{Interval} & \textbf{Adjacency type}\\
\hline
METR-LA & 207 & 34,272 & 5 min & Distance-based \\
PEMS-Bay & 325 & 52,116 & 5 min & Distance-based  \\
Chengdu & 524 & 17,280 & 10 min & Connectivity-based \\
Shenzhen & 627 & 17,280 & 10 min & Connectivity-based \\
\hline
\end{tabular}
\end{table}

The datasets span network sizes from 207 to 627 nodes and include both the US highway networks (METR-LA, PEMS-Bay) and Chinese urban networks (Chengdu, Shenzhen), enabling evaluation across diverse traffic patterns and network topologies. All datasets contain traffic speeds measured at sensors with a specific granularity. We use a consistent 70\%/10\%/20\% split for training, validation, and testing. We use a Z-score normalisation scheme to prepare the data.

\subsection{Experimental Setup}

\textbf{Task configuration.} We predict 12 future timesteps from 12 historical timesteps, corresponding to 1-hour prediction horizons for the 5-min datasets (METR-LA, PEMS-Bay) and 2-hour horizons for the 10-min datasets (Chengdu, Shenzhen). We report performance at 10-min, 30-min, and 60-min horizons, consistent with evaluation practices in literature \cite{SurveyModernDeeptedjopurnomo2022}. Performance is evaluated using mean absolute error, root mean squared error (RMSE), and mean absolute percentage error (MAPE).

\textbf{Model architectures.} Each ST-Conv block uses 2-hop Chebyshev polynomials for spatial modelling  ($K_s = 2$), temporal convolutions with kernel size $K_t = 3$ for the 1-block and 2-block variants, and $K_t = 2$ for the 3-block variant, as the temporal kernel size must be reduced in the deeper model to prevent complete exhaustion of the fixed input sequence length due to cumulative shrinkage across stacked ST-Conv blocks. The channel configurations are [1, 16, 64] for STGCN-1B, [1, 16, 64] $\rightarrow$ [64, 16, 64] for STGCN-2B, and [1, 16, 64] $\rightarrow$ [64, 16, 64] $\rightarrow$ [64, 16, 64] for STGCN-3B. Dropout is intentionally omitted in these variants to avoid introducing stochastic regularisation effects that could confound a controlled comparison of architectural depth.

\textbf{Training details.} Training configuration follows the implementation of \cite{STG4TrafficSurveyBenchmarkluo2024}. Models were trained using AdamW optimiser (lr=0.001, weight decay=0.0001) with a multi-step learning rate scheduling (decay 0.3 at epochs 60, 80, 100). The maximum number of training epochs was set to 100. We applied gradient clipping (maximum norm = 5.0) and early stopping with a patience of 50 epochs. Batch sizes were 64 for METR-LA and PEMS-Bay, and 32 for Chengdu and Shenzhen. The loss function was masked MAE computed on de-normalised predictions. All experiments were repeated across five random seeds; we report mean values.

\subsection{Computational Analysis}

To complement predictive performance, we evaluate each architecture along three computational dimensions: inference latency, throughput, and FLOPs. Inference latency measures the wall-clock time to process a single batch, throughput measures the number of predictions produced per second, and FLOPs quantify the number of floating-point operations performed during a single forward pass as a hardware-independent measure of computational complexity.

\textbf{Inference latency.} All measurements were conducted under CPU-only conditions to provide a hardware-agnostic baseline relevant to real-world deployment, using PyTorch \cite{paszke2019pytorch}. Each model was set to evaluation mode with gradient computation disabled via \texttt{torch.no\_grad()}. A warm-up pass was performed prior to timing to stabilise PyTorch's memory allocator, after which
$M=50$ batches were timed using \texttt{time.time()}, yielding 49 measured observations per run. This was repeated across five random seeds, with mean and standard deviation reported. Batch sizes were 64 for METR-LA and PEMS-Bay, and 32 for Chengdu and Shenzhen, reflecting their larger graphs.

\textbf{Throughput.} Throughput, $S$, was computed as total samples processed divided by total elapsed time across the 49 timed batches:
\begin{equation}
    S = \frac{\sum_{b=1}^{49} m_b}{\sum_{b=1}^{49} t_b},
\end{equation}
where $m_b$ and $t_b$ are the batch size and wall-clock duration of batch $b$, respectively. Per-seed throughput values were averaged across the five seeds to yield the reported figure.

\textbf{FLOPs.} Computational complexity was estimated using THOP \cite{zhu2018thop}, which traces PyTorch operations during a forward pass to approximate FLOPs. Profiling used a dummy input of shape $[1, 12, N, 1]$, representing a single sample with 12 timesteps and $N$ nodes, with Chebyshev polynomial matrices ($K_s = 2$) provided as auxiliary inputs. Results are reported in \Cref{tab:computational}.

Training was performed on NVIDIA RTX 5090 GPUs (32GB). CPU inference benchmarks were measured on an Intel Core Ultra 9 285K (24 cores, 5.1 GHz) using single-threaded execution for reproducibility.


\section{Results}
\subsection{Forecasting Performance}

\Cref{tab:forecasting} presents forecasting performance across all architecture variants and datasets. Several patterns emerge from the results:

\begin{table*}[t]
\centering
\caption{Forecasting performance comparison across 1-block, 2-block, and 3-block STGCN architectures (MAE, RMSE, MAPE) at 10/30/60 mins. Values are reported as mean $\pm$ std across seeds. The best results for each set are in bold.}
\label{tab:forecasting}

\begin{tabular}{llc|ccc}
\hline
Dataset & Horizon & Model & MAE & RMSE & MAPE (\%) \\
\hline

\multirow{9}{*}{METR-LA}
 & \multirow{3}{*}{10-min}
 & 1-block & 2.598 $\pm$ 0.001 & 4.862 $\pm$ 0.006 & 6.560 $\pm$ 0.030 \\
 &  & 2-block & \textbf{2.578 $\pm$ 0.005} & 4.807 $\pm$ 0.014 & 6.500 $\pm$ 0.020 \\
 &  & 3-block & 2.582 $\pm$ 0.013 & \textbf{4.782 $\pm$ 0.019} & \textbf{6.420 $\pm$ 0.060} \\
 \cline{2-6}
 & \multirow{3}{*}{30-min}
 & 1-block & 3.246 $\pm$ 0.005 & 6.607 $\pm$ 0.024 & 9.050 $\pm$ 0.100 \\
 &  & 2-block & 3.209 $\pm$ 0.011 & 6.514 $\pm$ 0.047 & 8.840 $\pm$ 0.040 \\
 &  & 3-block & \textbf{3.199 $\pm$ 0.021} & \textbf{6.466 $\pm$ 0.057} & \textbf{8.660 $\pm$ 0.080} \\
 \cline{2-6} 
 & \multirow{3}{*}{60-min}
 & 1-block & 3.757 $\pm$ 0.003 & 7.769 $\pm$ 0.020 & 11.020 $\pm$ 0.110 \\
 &  & 2-block & \textbf{3.706 $\pm$ 0.013} & 7.695 $\pm$ 0.048 & 10.720 $\pm$ 0.090 \\
 &  & 3-block & 3.717 $\pm$ 0.030 & \textbf{7.661 $\pm$ 0.067} & \textbf{10.550 $\pm$ 0.120} \\
\hline
\hline

\multirow{9}{*}{PEMS-Bay}
 & \multirow{3}{*}{10-min}
 & 1-block & \textbf{1.150 $\pm$ 0.002} & \textbf{2.282 $\pm$ 0.005} & \textbf{2.330 $\pm$ 0.010} \\
 &  & 2-block & 1.165 $\pm$ 0.009 & 2.300 $\pm$ 0.018 & 2.370 $\pm$ 0.020 \\
 &  & 3-block & 1.174 $\pm$ 0.003 & 2.316 $\pm$ 0.007 & 2.400 $\pm$ 0.020 \\
 \cline{2-6}
 & \multirow{3}{*}{30-min}
 & 1-block & \textbf{1.693 $\pm$ 0.002} & \textbf{3.855 $\pm$ 0.009} & \textbf{3.800 $\pm$ 0.020} \\
 &  & 2-block & 1.698 $\pm$ 0.009 & 3.865 $\pm$ 0.034 & 3.810 $\pm$ 0.030 \\
 &  & 3-block & 1.708 $\pm$ 0.011 & 3.899 $\pm$ 0.040 & 3.850 $\pm$ 0.060 \\
 \cline{2-6}
 & \multirow{3}{*}{60-min}
 & 1-block & \textbf{2.039 $\pm$ 0.007} & \textbf{4.706 $\pm$ 0.021} & 4.790 $\pm$ 0.030 \\
 &  & 2-block & 2.043 $\pm$ 0.011 & 4.736 $\pm$ 0.029 & 4.810 $\pm$ 0.060 \\
 &  & 3-block & 2.049 $\pm$ 0.022 & 4.755 $\pm$ 0.054 & \textbf{4.780 $\pm$ 0.070} \\
\hline
\hline


\multirow{9}{*}{Chengdu}
 & \multirow{3}{*}{10-min}
 & 1-block & \textbf{1.947 $\pm$ 0.007} & \textbf{2.872 $\pm$ 0.009} & \textbf{8.380 $\pm$ 0.030} \\
 &  & 2-block & 1.957 $\pm$ 0.008 & 2.890 $\pm$ 0.009 & 8.470 $\pm$ 0.030 \\
 &  & 3-block & 1.974 $\pm$ 0.003 & 2.916 $\pm$ 0.007 & 8.560 $\pm$ 0.050 \\
 \cline{2-6}
 & \multirow{3}{*}{30-min}
 & 1-block & 2.282 $\pm$ 0.013 & 3.457 $\pm$ 0.019 & 10.350 $\pm$ 0.060 \\
 &  & 2-block & 2.260 $\pm$ 0.011 & 3.425 $\pm$ 0.018 & 10.230 $\pm$ 0.070 \\
 &  & 3-block & \textbf{2.255 $\pm$ 0.008} & \textbf{3.415 $\pm$ 0.014} & \textbf{10.180 $\pm$ 0.100} \\
 \cline{2-6}
 & \multirow{3}{*}{60-min}
 & 1-block & 2.412 $\pm$ 0.012 & 3.674 $\pm$ 0.018 & 11.140 $\pm$ 0.070 \\
 &  & 2-block & 2.389 $\pm$ 0.023 & 3.645 $\pm$ 0.044 & 10.970 $\pm$ 0.160 \\
 &  & 3-block & \textbf{2.366 $\pm$ 0.016} & \textbf{3.605 $\pm$ 0.026} & \textbf{10.820 $\pm$ 0.120} \\
\hline
\hline


\multirow{9}{*}{Shenzhen}
 & \multirow{3}{*}{10-min}
 & 1-block & \textbf{1.702 $\pm$ 0.005} & \textbf{2.561 $\pm$ 0.008} & \textbf{7.020 $\pm$ 0.030} \\
 &  & 2-block & 1.708 $\pm$ 0.005 & 2.581 $\pm$ 0.013 & 7.090 $\pm$ 0.030 \\
 &  & 3-block & 1.722 $\pm$ 0.006 & 2.602 $\pm$ 0.013 & 7.130 $\pm$ 0.060 \\
 \cline{2-6}
 & \multirow{3}{*}{30-min}
 & 1-block & 2.010 $\pm$ 0.005 & 3.145 $\pm$ 0.011 & 8.570 $\pm$ 0.040 \\
 &  & 2-block & 1.986 $\pm$ 0.004 & 3.112 $\pm$ 0.011 & 8.460 $\pm$ 0.030 \\
 &  & 3-block & \textbf{1.984 $\pm$ 0.010} & \textbf{3.110 $\pm$ 0.020} & \textbf{8.430 $\pm$ 0.070} \\
 \cline{2-6}
 & \multirow{3}{*}{60-min}
 & 1-block & 2.147 $\pm$ 0.008 & 3.435 $\pm$ 0.016 & 9.280 $\pm$ 0.050 \\
 &  & 2-block & 2.109 $\pm$ 0.007 & 3.376 $\pm$ 0.017 & 9.090 $\pm$ 0.060 \\
 &  & 3-block & \textbf{2.098 $\pm$ 0.017} & \textbf{3.358 $\pm$ 0.030} & \textbf{8.990 $\pm$ 0.080} \\
\hline
\hline

\end{tabular}
\end{table*}

\textbf{Diminishing returns from depth.} Across all datasets, the performance gap between architectures is small. The transition from 1-block to 2-block yields modest improvements at longer horizons: 1.0–1.8\% MAE reduction at 60-min for three datasets (METR-LA, Chengdu, Shenzhen). However, the 3-block variant provides negligible additional gains over 2-block ($<$0.5\% MAE improvement at any horizon), suggesting the 2-block architecture already saturates STGCN's representational capacity.

\textbf{Horizon-dependent patterns.} The 1-block variant outperforms deeper variants at short-term horizons (10 mins) for three of four datasets (PEMS-Bay, Chengdu, Shenzhen). The benefit of depth only materialises at longer horizons (30–60 mins). This suggests that additional spatio-temporal blocks primarily help capture longer-range temporal dependencies, while short-term prediction is adequately served by a single block.

\textbf{Dataset-dependent patterns.} PEMS-Bay represents an extreme case: 1-block outperforms both 2-block and 3-block across nearly all horizons and metrics, consistent across all initialisations. This indicates that additional depth can be counterproductive, depending on the dataset's characteristics.

\begin{table*}[t]
\centering
\caption{Computational comparison across STGCN depth variants. Percentages indicate an increase relative to 1-block. CPU inference times are shown as mean $\pm$ std format across individual runs.}
\label{tab:computational}

\begin{tabular}{llllllc}

\hline
\textbf{Dataset} & \textbf{Model} & \textbf{Params} & \textbf{MFLOPs} & \textbf{Train (s)} &  \textbf{Latency (ms)} & \textbf{Throughput (pred/s)} \\
\hline
 & 1-block & 127,388 & 20.34 & 142.9 & 216.2 $\pm$ 3.7 & 296 \\
METR-LA & 2-block & 131,996 (+3.6\%) & 25.63 (+26.0\%) & 201.9 (+41.3\%) & 347.1 $\pm$ 11.8 (+60.5\%) & 185 \\
& 3-block & 178,556 (+40.2\%) & 40.21 (+97.7\%) & 273.5 (+91.4\%) & 503.1 $\pm$ 4.7 (+132.7\%) & 127 \\
\hline
& 1-block & 157,596 & 31.93 & 300.3 & 229.3 $\pm$ 4.1 & 279 \\
PEMS-Bay & 2-block & 177,308 (+12.5\%) & 40.25 (+26.1\%) & 435.0 (+44.9\%) & 369.9 $\pm$ 5.3 (+61.3\%) & 173 \\
& 3-block & 238,972 (+51.6\%) & 63.13 (+97.7\%) & 711.4 (+136.9\%) & 535.9 $\pm$ 2.9 (+133.7\%) & 119 \\
\hline
& 1-block & 208,540 & 51.48 & 363.2 & 230.3 $\pm$ 1.7 & 139 \\
Chengdu & 2-block & 253,724 (+21.7\%) & 64.89 (+26.0\%) & 427.4 (+17.7\%) & 371.6 $\pm$ 3.9 (+61.4\%) & 86 \\
& 3-block & 340,860 (+63.5\%) & 101.78 (+97.7\%) & 452.0 (+24.4\%) & 541.5 $\pm$ 6.1 (+135.1\%) & 59\\
\hline
& 1-block & 234,908 & 61.60 & 430.8 & 235.6 $\pm$ 3.1 & 136 \\
Shenzhen & 2-block & 293,276 (+24.8\%) & 77.65 (+26.1\%) & 511.5 (+18.7\%) & 379.4 $\pm$ 8.4 (+61.0\%) & 84 \\
& 3-block & 393,596 (+67.6\%) & 121.79 (+97.7\%) & 544.0 (+26.3\%) & 551.3 $\pm$ 3.0 (+134.0\%) & 58 \\
\hline
\end{tabular}
\end{table*}

\subsection{Computational Efficiency}

\Cref{tab:computational} presents computational metrics across all architecture variants. All percentage comparisons are reported relative to the 1-block baseline.

\textbf{FLOPs scaling.} FLOPs increase consistently across all datasets: the 2-block variant requires approximately 26\% more FLOPs than the 1-block, while the 3-block variant requires approximately 98\% more -- nearly double. This near-linear scaling reflects the sequential nature of the spatio-temporal blocks.

\textbf{Parameter scaling.} Parameter count grows modestly from 1-block to 2-block (+3.6\% to +24.8\% depending on dataset) but more substantially to 3-block (+40.2\% to +67.6\%). The relatively small increase from 1-block to 2-block is due to a large proportion of parameters residing in the output projection layer, which is shared across variants.

\textbf{Training time.} Training time increases by 18-45\% from 1-block to 2-block, with larger relative increases on the highway datasets (METR-LA: +41.3\%, PEMS-Bay: +44.9\%) compared to the urban datasets (Chengdu: +17.7\%, Shenzhen: +18.7\%).

\subsection{CPU Inference Analysis}

For deployment-relevant evaluation, we measure CPU inference latency and throughput, as many real-world ITS deployments operate on CPU-based systems due to cost, power, and infrastructure constraints.

\textbf{Latency.} The $\sim$61\% latency increase from 1-block to 2-block contrasts with only a $\sim$26\% increase in FLOPs, reflecting per-block dispatch and memory overheads inherent to CPU inference rather than a measurement artefact. The 2-block variant incurs approximately 61\% higher latency than the 1-block baseline across all datasets (METR-LA: +60.5\%, PEMS-Bay: +61.3\%, Chengdu: +61.4\%, Shenzhen: +61.0\%). This consistency across network sizes ranging from 207 to 627 nodes suggests the finding generalises reliably. The 3-block variant incurs over double the latency of 1-block across all datasets (METR-LA: +132.7\%, PEMS-Bay: +133.7\%, Chengdu: +135.1\%, Shenzhen: +134.0\%). The combined efficiency and accuracy results are summarised in Fig. 1.

\textbf{Throughput.} The 1-block variant sustains approximately 60\% higher throughput than 2-block across all datasets: 296 vs.\ 185 pred/s on METR-LA (+60.0\%), 279 vs.\ 173 on PEMS-Bay (+61.3\%), 139 vs.\ 86 on Chengdu (+61.6\%), and 136 vs.\ 84 on Shenzhen (+61.9\%). The substantially lower absolute throughput on Chengdu and Shenzhen compared to METR-LA and PEMS-Bay reflects the greater computational cost of the Chebyshev graph convolution as the number of nodes increases from 207/325 to 524/627.

\subsection{Efficiency-Performance Tradeoff}

Fig. 1 visualises the efficiency–accuracy tradeoff across all four datasets, with both axes normalised to the 1-block baseline. Points in the lower-left shaded region are Pareto-optimal -- simultaneously achieving lower latency and lower MAE. On PEMS-Bay, the 1-block variant is the sole Pareto-optimal point; deeper variants move strictly away from the optimum, incurring higher latency while also degrading accuracy. On the remaining three datasets, 2-block and 3-block variants shift substantially rightward (1.6× and 2.3× latency, respectively) for accuracy improvements that are negligible in absolute terms -- note that the compressed y-axis reflects the true scale of these differences. In no dataset does the 3-block variant offer a favourable tradeoff over 2-block, let alone 1-block.

\section{Discussion}

\textbf{Implications for deployment.} The 1-block variant presents 
a compelling case for practical deployment. It reduces CPU inference 
latency by approximately 38\% and increases throughput by approximately 
60\% relative to the 2-block baseline, while incurring at most 1.8\% 
MAE degradation at longer horizons and matching or outperforming 2-block 
at shorter horizons. For resource-constrained or high-frequency deployment 
scenarios -- such as adaptive signal control operating at city scale -- 
this tradeoff strongly favours the shallower architecture. 

\textbf{Horizon-dependent value of depth.} Our results reveal that 
architectural depth interacts with prediction horizon in a consistent 
and practically meaningful way. For short-term forecasting (10–15 mins), 
which is most relevant for real-time signal control and immediate routing 
decisions, 1-block is optimal or near-optimal across all datasets. The 
benefit of additional blocks only materialises at longer horizons (30-60 
mins), where deeper architectures appear better able to capture extended 
temporal dependencies. Crucially, longer-horizon applications also tend 
to be more tolerant of latency -- a planning system operating on 60-min forecasts is less sensitive to inference time than those driving real-time 
signal actuation. The result is that depth provides marginal accuracy 
benefit precisely in the scenarios where its computational cost matters 
least, and provides no benefit in the scenarios where efficiency is most 
critical.

\textbf{Dataset characteristics and depth sensitivity.} The PEMS-Bay result warrants particular attention. Across all horizons and metrics, 
1-block not only matches but consistently outperforms deeper variants, 
with the gap widening rather than closing at longer horizons. This 
suggests that additional spatio-temporal blocks can be counterproductive 
depending on dataset characteristics -- potentially introducing 
overfitting or over-smoothing in networks where the spatial or temporal structure is more regular. The US highway network captured in PEMS-Bay 
may present a more structured and predictable signal than the Chinese 
urban networks, making it less reliant on the additional representational 
capacity that deeper blocks provide. Understanding what dataset properties 
drive this sensitivity is an open question, but the result reinforces 
that the 2-block default should not be treated as universally appropriate.

\textbf{Implications for benchmarking.} Beyond deployment, our findings 
carry methodological implications for the research community. Numerous 
efficiency-focused works benchmark against the standard 2-block STGCN 
as their GCN-based baseline for computational comparisons 
\cite{liu2023we, jiang2024stgformer, Zhang_Gao_Wang_Yiu_Yin_2025}. 
Since 1-block STGCN matches or exceeds 2-block accuracy on the majority 
of datasets at 38\% lower inference latency, these comparisons may 
overstate the efficiency gains of proposed methods relative to the minimal 
competitive architecture. We suggest that the 1-block STGCN is a more 
appropriate efficiency baseline -- one that better reflects the true 
lower bound of what a graph-based spatio-temporal model needs to be 
competitive, and that sets a higher bar for methods claiming computational 
improvements.

\textbf{Broader architectural questions.} Our findings raise a question 
that extends beyond STGCN: are stacked designs in other STGNNs similarly 
over-parameterised? Models such as DCRNN and Graph WaveNet adopt 
analogous depth conventions that have been inherited across the literature 
without systematic ablation. The mechanism underlying our result -- that 
a large proportion of parameters reside in the shared output projection 
layer, meaning additional blocks contribute relatively little to the parameter 
count but substantially to inference cost -- may apply more broadly to 
architectures with similar designs. Whether the depth-performance patterns 
observed here generalise to these models is an open and practically 
important question.

\section{Limitations}
While this work systematically analyses architectural depth in STGCN and finds that a lighter model performs comparably to the standard benchmark, the analysis is specific to the STGCN architecture. Other spatio-temporal graph networks, such as DCRNN and Graph WaveNet, may exhibit distinct depth-performance patterns and warrant separate investigation.

Further, although four geographically diverse datasets are examined to support generalisation, all datasets capture traffic speed. Extending the analysis to traffic flow and occupancy datasets where STGCN is prevalently used is a natural next step.

Finally, CPU inference benchmarks were conducted in single-threaded mode on a high-end processor, chosen to approximate a controlled, reproducible deployment baseline. Absolute latency figures will differ on lower-end or embedded hardware; however, the relative differences between architecture variants are determined by model complexity rather than hardware capability, and are expected to remain consistent across CPU configurations.

\section{Conclusion}

We conducted a systematic analysis of STGCN architectural depth for traffic prediction, comparing 1-block, 2-block, and 3-block variants across four diverse datasets. Our findings reveal that the single-block variant is optimal for short-term prediction (10-15 mins) on three of four datasets, while incurring only marginal degradation ($\leq$1.8\% MAE) at longer horizons. Compared with the standard 2-block architecture, the 1-block variant reduces CPU inference latency by approximately 38\% and increases throughput by approximately 60\% -- substantial gains for resource-constrained ITS deployments. The 3-block architecture offers no favourable tradeoff in any scenario, more than doubling inference cost for negligible or negative predictive return.

These results suggest that the default 2-block STGCN may be over-parameterised for many practical applications, with implications for both practitioners selecting models for deployment and researchers establishing efficiency baselines. For real-time ITS applications requiring short-term forecasts, we recommend the 1-block STGCN as the default architectural choice.

\bibliographystyle{IEEEtran}
\bibliography{references}

\end{document}